\pdfoutput=1

\documentclass{article}
\usepackage{colt09e,times}
\usepackage[scaled]{helvet}
\usepackage[british]{babel}
\usepackage{graphicx}
\usepackage{amsmath,amsfonts,amsthm,latexsym}
\usepackage{subfig}
\usepackage{float}
\usepackage{algorithm,algorithmicx,algpseudocode}
\usepackage{mathtools,commath}
\usepackage[all]{xy}
\usepackage{booktabs}
\usepackage{wdel}
\usepackage{soul} %
\usepackage{stmaryrd} %
\usepackage{enumerate}
\usepackage{nicefrac}
\usepackage{paralist}   %
\usepackage{afterpage}

\newxyColor{myshadea}{.7}{gray}{}
\newxyColor{myshadeb}{.5}{gray}{}
\newxyColor{myshadec}{.3}{gray}{}
\newxyColor{myshaded}{.1}{gray}{}

\floatstyle{ruled}
\restylefloat{figure}
\restylefloat{table}
\restylefloat{algorithm}

\newtheorem{theorem}{Theorem}
\newtheorem{corollary}{Corollary}
\newtheorem{lemma}{Lemma}
\theoremstyle{definition}

\newtheorem{exampleX}{Example}[section] %

\newenvironment{example}[1][nope]{%
  \ifthenelse{\equal{#1}{nope}}{\begin{exampleX}}{\begin{exampleX}[#1]}%
  \pushQED{\qed}%
}{%
  \popQED%
  \end{exampleX}%
}

\let\del\undefined
\let\set\undefined
\let\sbr\undefined

\DeclarePairedDelimiter\del{\lparen}{\rparen}
\DeclareTripledDelimiter\delc{\lparen}{\vert}{\rparen}{\mathord}
\DeclareTripledDelimiter\delcc{\lparen}{\Vert}{\rparen}{\mathord}
\DeclarePairedDelimiter\set{\lbrace}{\rbrace}
\DeclareTripledDelimiter\setc{\lbrace}{\vert}{\rbrace}{\mathrel}
\DeclarePairedDelimiter\sbr{\lbrack}{\rbrack}
\DeclareTripledDelimiter\sbrc{\lbrack}{\vert}{\rbrack}{\mathord}
\DeclarePairedDelimiter\tuple{\langle}{\rangle}
\DeclarePairedDelimiter\card{\lvert}{\rvert}

\DeclareMathSymbol{:}{\mathord}{operators}{`:}
\newcommand{\df}{\vcentcolon\nolinebreak\mkern-1.2mu=}

\newcommand{\algoref}[1]{Algorithm~\ref{#1}}

\newcommand{\corlref}[1]{Corollary~\ref{#1}}
\newcommand{\tabref}[1]{Table~\ref{#1}}
\newcommand{\markdef}[1]{\emph{#1}} %
\newcommand{\family}[3][-1]{\tuple[#1]{#2}_{#3}}  %

\newcommand{\OurAlgNotFancy}{EPP}
\newcommand{\algname}[1]{\textnormal{\textsc{#1}}}
\newcommand{\OurAlg}{\algname{\OurAlgNotFancy}}
\newcommand{\FreezAlg}{\OurAlg\algname{-Freezing}}
\newcommand{\SleepAlg}{\OurAlg\algname{-Sleeping}}

\newcommand{\sups}[1][nope]{%
\ifthenelse{\equal{#1}{nope}}{}{^{#1}}}
\newcommand{\subs}[1][nope]{%
\ifthenelse{\equal{#1}{nope}}{}{_{#1}}}

\newcommand{\notationbuddy}[2]{%
  \uppercase{\def\thisupper{#1}}%
  \expandafter\newcommand\csname\thisupper\endcsname{\mathcal{\uppercase{#2}}}
  \expandafter\newcommand\csname#1\endcsname{#2}
  \expandafter\newcommand\csname#1s\endcsname[1][1:T]{{\csname#1\endcsname}_{##1}}
  \expandafter\newcommand\csname#1rv\endcsname{\uppercase{#2}}
  \expandafter\newcommand\csname\thisupper s\endcsname[1][1:T]{{\csname#1rv\endcsname}_{##1}}
}

\let\a\undefined
\notationbuddy{x}{x} %
\notationbuddy{a}{a} %
\notationbuddy{e}{e} %
\notationbuddy{q}{q} %

\renewcommand{\xi}{\e}         %
\renewcommand{\Xi}{\E}         %
\newcommand{\Xis}{\Es}         %
\let\actions\A                 %
\let\expert\e                  %
\let\Expert\erv                %
\let\expertset\E               %

\newcommand{\freezcomp}[1][\Cell]{\A^\fr_{#1}}  %
\newcommand{\sleepcomp}[1][\Cell]{\A^\sl_{#1}}  %
\newcommand{\foscomp}[1][\Cell]{\A^\fos_{#1}}   %

\newcommand{\N}{\mathbb N}                        %
\DeclareMathOperator{\loss}{\ell}                    %
\newcommand{\logloss}{{\ell\ell}}
\DeclareMathOperator{\Spmass}{p}                  %
\DeclareMathOperator{\Spfs}{P}                    %
\DeclareMathOperator*{\ex}{E}                     %

\DeclareMathOperator*{\Union}{\bigcup}

\newcommand{\half}{\ensuremath{\nicefrac{1}{2}}}

\newcommand{\pmass}{{\Spmass}\subs}
\newcommand{\Pmass}{{\Spfs}\subs}

\newcommand{\hmml}[1]{\operatorname{#1}}             %

\newcommand{\init}[1]{\hmml{#1}_\circ}               %
\newcommand{\tf}[1]{\hmml{#1}_\shortrightarrow\sups} %
\newcommand{\pf}[1]{\hmml{#1}_\shortdownarrow\sups}  %

\newcommand{\pinit}{\init{p}}              %
\newcommand{\ptf}{\tf{p}}                  %
\newcommand{\ppf}{\pf{p}}                  %

\newcommand{\ntransitions}{g}             %
\newcommand{\hmm}[1]{\mathfrak{#1}}
\renewcommand{\A}{\hmm{H}}                          %

\newcommand{\modifier}[1]{\textnormal{\sffamily #1}}
\renewcommand{\sl}{\modifier{sl}}
\newcommand{\fr}{\modifier{fr}}
\newcommand{\fos}{\modifier{f/s}}%

\newcommand{\Part}{\mathbb C}                     %
\newcommand{\Cell}{\mathcal C}                    %
\newcommand{\prev}{\operatorname{prev}}

\newcommand{\Alg}{\modifier{alg}}
\newcommand{\AlgAlg}{\algname{Alg}}

\newcommand{\PastPost}{\pi}          %
\newcommand{\PredPost}{\lambda}      %
\newcommand{\Joint}{\MyPred_t}       %
\newcommand{\BackRef}{\beta}         %

\newcommand{\Pred}{\pmass\sups}
\newcommand{\ExPred}[1]{\Pred[#1]}             %
\newcommand{\MyPred}{\Pred[\Alg]}              %
\newcommand{\ExPreds}{\Pred[\Xi]}              %

\newcommand{\Act}{\a\sups}
\newcommand{\ExAct}[1]{\Act[#1]}              %
\newcommand{\MyAct}{\Act[\Alg]}               %
\newcommand{\ExActs}{\Act[\Xi]}               %

\newcommand{\mpp}{\modifier{MPP}}
\newcommand{\bayes}{\modifier{B}}

\newcommand{\mppprob}{\Pmass^\mpp\subs}
\newcommand{\bayesprob}{\Pmass^\bayes\subs}
\newcommand{\frprob}{\Pmass^\fr\subs}
\newcommand{\slprob}{\Pmass^\sl\subs}
\newcommand{\fosprob}{\Pmass^\fos\subs}

\newcommand{\rv}[1]{*+=+[o][F]{#1}}     %
\newcommand{\lt}[2]{#1^*{#2}}           %

\newcommand{\exclude}[1]{}

\title{Freezing and Sleeping:\\Tracking Experts that Learn by Evolving
Past Posteriors}
\author{Wouter M. Koolen \and Tim van Erven\thanks{%
We would like to thank Manfred Warmuth for raising our interest in this
subject during COLT 2008. We also thank Steven de Rooij and Peter
Gr\"unwald for fruitful discussions and suggestions. This work was
supported in part by the IST Programme of the European Community, under
the PASCAL Network of Excellence, IST-2002-506778. This publication only
reflects the authors' views.}\\
Centrum Wiskunde \& Informatica (CWI)\\
Science Park 123, P.O. Box 94079 \\
1090 GB Amsterdam, The Netherlands\\
{\tt \{Wouter.Koolen,Tim.van.Erven\}@cwi.nl}}

\begin{document}
\exclude{
\pagenumbering{roman}  %
\tableofcontents
\cleardoublepage
\pagenumbering{arabic} %
}
\maketitle

\begin{abstract}
A problem posed by Freund is how to efficiently track a small pool of
experts out of a much larger set. This problem was solved when Bousquet and
Warmuth introduced their mixing past posteriors (MPP) algorithm in 2001.

In Freund's problem the experts would normally be considered black
boxes. However, in this paper we re-examine Freund's problem in case the
experts have internal structure that enables them to learn. In this case
the problem has two possible interpretations: should the experts learn
from all data or only from the subsequence on which they are being
tracked? The MPP algorithm solves the first case. Our contribution is to
generalise MPP to address the second option. The results we obtain apply
to any expert structure that can be formalised using (expert) hidden
Markov models. Curiously enough, for our interpretation there are \emph{two}
natural reference schemes: freezing and sleeping. For each scheme, we
provide an efficient prediction strategy and prove the relevant loss
bound. \end{abstract}

\section{Introduction}

Freund's problem arises in the context of prediction with expert
advice~\cite{cesa-bianchi2006}. In this setting a sequence of outcomes
needs to be predicted, one outcome at a time. Thus, prediction proceeds
in rounds: in each round we first consult a set of experts, who give us
their predictions. Then we make our own prediction and incur some loss
based on the discrepancy between this prediction and the actual outcome.
The goal is to minimise the difference between our cumulative loss and
some reference scheme. For this reference there are several options; we
may, for example, compare ourselves to the cumulative loss of the best
expert in hindsight. A more ambitious reference scheme was proposed by
Yoav Freund in 2000.

\paragraph{Freund's Problem} Freund asked for an efficient prediction
strategy that suffers low additional loss compared to the following
reference scheme:
\begin{enumerate}[(a)]
\setlength{\itemsep}{0pt}
\setlength{\parskip}{0pt}
\item\label{it:partition.data}
  Partition the data into several subsequences.
\item \label{it:choose.experts}
  Select an expert for each subsequence.
\item \label{it:sum.expert.loss}
  Sum the loss of the selected experts on their subsequences.
\end{enumerate}
\noindent
In 2001, Freund's problem was addressed by Bousquet and Warmuth, who
developed the efficient mixing past posteriors (MPP)
algorithm~\cite{bousquet2002}. MPP's loss is bounded by the loss of
Freund's scheme plus some overhead that depends on the number of bits
required to encode the partition of the data, and it has found
successful application in~\cite{Gramacy02adaptivecaching}. Problem
solved. Or is it?

\paragraph{The Loss of an Expert on a Subsequence}
In our view Freund's problem has two possible interpretations, which
differ most clearly for learning experts.
Namely, to measure the predictive performance of an expert on a
subsequence, do we show her the data \emph{outside} her subsequence or
not? An expert that sees all outcomes will track the \emph{global}
properties of the data. This is (implicitly) the case for mixing past
posteriors. But an expert that only observes the subsequence that she
has to predict might see and thus exploit its \emph{local} structure,
resulting in decreased loss. The more the characteristics of the
subsequences differ, the greater the gain. Let us illustrate this by an
example.

\paragraph{Ambiguity Example} The data consist of a block of ones,
followed by a block of zeros, again followed by a block of ones. In the reference scheme
step~(\ref{it:partition.data}), we split the data into two subsequences,
one consisting of only ones, the other consisting of only zeros. Our
expert predicts the probability of a one using Laplace's rule of
succession, i.e.\ she \emph{learns} the frequency of ones in the data
that she observes~\cite{cesa-bianchi2006}. Note that one learning expert
suffices, as we can select her for both subsequences.

\begin{figure}
\caption{Estimated probability of a one}\label{fig:example}
\centering
\includegraphics[trim= .8cm 0 .2cm 0]{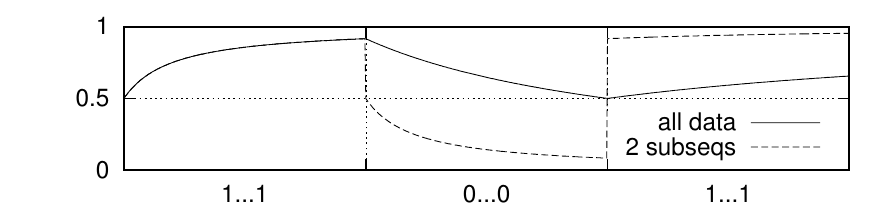}
\end{figure}

First we consider the expert's predictions when she observes all data. During the first block, our expert will increase her probability of a one from $\half$ to nearly one. Then during the second block it will go down to $\half$ again. During the third block it will increase from $\half$ up, but slower. Thus, for block two the expert is extremely bad, while for block three she is at best mediocre. (See
\figref{fig:example}.)

Compare this to the expert's predictions on the subsequences. During the subsequence of ones (first and third block), our expert will increase her probability of a one from $\half$ to almost one, while during the subsequence of zeroes she will decrease her probability from $\half$ to nearly zero. Thus, the expert is much better on the subsequences in isolation.

This shows that the predictive performance of a learning expert on a subsequence in isolation can be dramatically higher than that on the same sequence in the context of all data. This behaviour is typical: on all data a learning expert will learn the average, global pattern, while on a well-chosen subsequence she can zoom in on the local structure.

\paragraph{Structured Experts}
In this paper, we solve Freund's problem under the interpretation that experts only observe the subsequence on which they are evaluated. Of course, for \emph{arbitrary} experts, this is impossible. For in the setting of prediction with experts, the expert predictions that we receive each round are \emph{always} in the context of all data. We have no access to the experts' predictions in the context of any subsequence, which may differ drastically from those on the whole data.

Often however, experts have internal structure. For example,
in~\cite{LittlestoneWarmuth1994,HerbsterWarmuth1998,volfwillems1998,Vovk1990}
adaptive prediction strategies (i.e., learning experts) are explicitly
constructed from basic experts. To represent such structured experts, we
use a general framework called \emph{expert hidden Markov models}
(EHMMs), that was introduced
in~\cite{koolen08:_combin_exper_advic_effic}. %
EHMMs are hidden Markov models in which the production probabilities are
determined by expert advice.  A structured expert in EHMM form provides
sufficient information about its predictions on
any isolated subsequence.

Many strategies for prediction with expert advice (i.e.\ learning
experts) can be rendered as EHMMs. For example all adaptive strategies in
the papers above (see~\cite{koolen08:_combin_exper_advic_effic}). But
there are also strategies that cannot be brought into EHMM form, like
e.g.\ \emph{follow the perturbed leader}~\cite{hannan57} and
\emph{variable share}~\cite{HerbsterWarmuth1998}.

Our approach may also be of interest to machine learning with regular
hidden Markov models (HMMs)~\cite{rabiner1989}. Although existing
approaches to shift between multiple HMMs~\cite{ghahramani00variational,ghahramani97:_factor_hidden_markov_model,landwehr08:_model_inter_hidden_proces} usually focus on change-point detection, prediction seems a highly related issue.

\paragraph{Sleeping or Freezing}
We evaluate the performance of learning experts on subsequences in
isolation. But now another choice presents itself (see
\figref{fig:mot.ex}). Should we present the subsequence to the expert
consecutively (we view this as \emph{freezing} the expert's state on
other data)? Or should we retain the original timing of the selected
samples and keep the intermediate samples unobserved (then the expert is
\emph{sleeping} for other data)? 
To illustrate the difference, consider an expert that is able to predict the television images of our favourite show. We want to \emph{freeze} her during commercial breaks, so that she continues predicting the show where it left off. We want to put her to \emph{sleep} when we zap to another channel, so that after zapping back, she will predict the show as it has advanced. 
Thus freezing vs sleeping is a modelling decision, that should be made on a case-by-case basis. We cover both scenarios.
\newenvironment{datatabular}[1]{\setlength{\tabcolsep}{.475em}\begin{tabular}{#1}}{\end{tabular}}
\begin{figure}
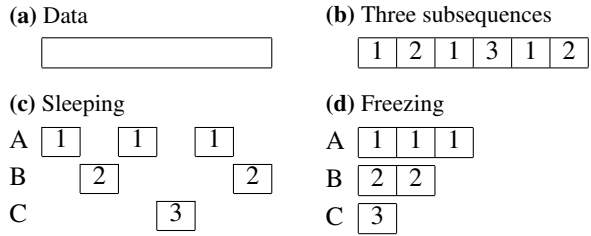

\caption{Freezing (consecutive) and Sleeping (timing preserved). Both use experts A,B,C on 
 subsequences 1,2,3. \label{fig:mot.ex}}
\centering
\begin{tabular}{ll}
\subfloat[Data\label{fig:mot.ex.data}]{%
\begin{tabular}{@{}l@{~~}c}
\phantom{A}&
\begin{datatabular}{|cccccc|}
\hline
\phantom{1} & \phantom{2} & \phantom{1} & \phantom{3} & \phantom{1} & \phantom{2} \\
\hline
\end{datatabular}
\end{tabular}
}
&
\subfloat[Three subsequences\label{fig:mot.ex.partition}]{%
\begin{tabular}{@{}l@{~~}c}
\phantom{A}&
\begin{datatabular}{|c|c|c|c|c|c|}
\hline
1 & 2 & 1 & 3 & 1 & 2 \\
\hline
\end{datatabular}
\end{tabular}
}
\\
\subfloat[Sleeping\label{fig:mot.ex.individual}]{%
\begin{tabular}{@{}l@{~~}c}
A& \begin{datatabular}{|c|c|c|c|c|c}
\cline{1-1}\cline{3-3}\cline{5-5}
1 & \phantom{2} & 1 & \phantom{3} & 1 & \phantom{2} \\
\cline{1-1}\cline{3-3}\cline{5-5}
\end{datatabular}
\\[.6ex]
B& \begin{datatabular}{c|c|ccc|c|}
\cline{2-2}\cline{6-6}
\phantom{1} & 2 & \phantom{1} & \phantom{3} & \phantom{1} & 2 \\
\cline{2-2}\cline{6-6}
\end{datatabular}
\\[.6ex]
C& \begin{datatabular}{ccc|c|cc}
\cline{4-4}
\phantom{1} & \phantom{2} & \phantom{1} & 3 & \phantom{1} & \phantom{2} \\
\cline{4-4}
\end{datatabular}
\end{tabular}%
}
&
\subfloat[Freezing\label{fig:mot.ex.consecutive}]{%
\begin{tabular}{@{}l@{~~}l}
A& \begin{datatabular}{|c|c|c|}
\hline
1 & 1 & 1 \\
\hline
\end{datatabular}
\\[.6ex]
B& \begin{datatabular}{|c|c|}
\hline
2 & 2\\
\hline
\end{datatabular}
\\[.6ex]
C& \begin{datatabular}{|c|}
\hline
3 \\
\hline
\end{datatabular}%
\end{tabular}
}%
\end{tabular}
\end{figure}

\subsection{Overview}

After preliminaries we start by reviewing the main existing loss bound
for mixing past posteriors in \secref{sec:mpp.intro}. Then, in
\secref{sec:ehmm}, we introduce EHMMs as a way to represent structured
experts.

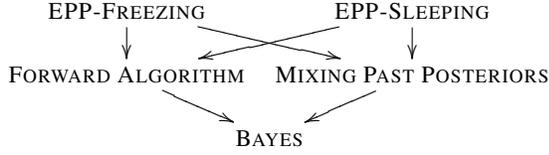
\begin{figure}
\caption{Generalisation relation among prediction strategies}\label{fig:gen.rel}
\centering
\footnotesize
$\xymatrix@!0@C=6em{
\FreezAlg \ar[drr] \ar[d] && \SleepAlg \ar[d] \ar[dll]
\\
\algname{Forward Algorithm} \ar[dr] && \algname{Mixing Past Posteriors} \ar[dl]
\\
& \algname{Bayes} &
}$
\end{figure}

The next section, \secref{sec:freezingandsleeping}, contains our results
for Freund's problem when structured experts are evaluated on
isolated subsequences. We formalise sleeping and freezing
as two different ways of presenting a subsequence of the data to an
EHMM, and present the \emph{evolving past posteriors} \OurAlg{}
algorithm that takes an EHMM as input. The \OurAlg{} algorithm has two
variants, which both generalise the mixing past posteriors algorithm in
a different way: \SleepAlg{} for sleeping and \FreezAlg{} for freezing.
The relation between \OurAlg{} and other existing prediction strategies
is shown in \figref{fig:gen.rel}. There $A \to B$ means that by
carefully choosing prediction strategy $A$'s parameters it reduces to
strategy $B$.

In order to understand \OurAlg{}, we verify that it produces the same
predictions for any two EHMMs that are equivalent in an appropriate
sense, and analyse its running time. We then proceed to show our main
result, which is that the losses of \FreezAlg{} and \SleepAlg{} are
bounded by the loss of Freund's scheme plus a complexity penalty that
depends on the number of bits required to encode the reference partition
in the same way as for mixing past posteriors. In fact, our bounds
(slightly) improve the known loss bound for mixing past posteriors.
Thus we solve Freund's problem with learning experts presented as EHMMs,
both for freezing and for sleeping.

We first derive our results only for logarithmic loss. This allows us to
use familiar concepts and results from probability theory and refer to
the interpretation of log loss as a codelength~\cite{cesa-bianchi2006}.
In \secref{sec:mixable.losses} we conclude by proving that any algorithm
that satisfies certain weak conditions, in particular \OurAlg{},
directly generalises to an algorithm for arbitrary mixable losses with
the appropriate loss bounds.

\section{Preliminaries}\label{sec:preliminaries}

\paragraph{Prediction With Expert Advice}\label{sec:expertadvice}
Each round $t$, we first receive advice from each expert $\xi \in \Xi$
in the form of an action $\as[t]^{\xi} \in \actions$. Then we distill our
own action $\MyAct_t \in \actions$ from the expert advice. Finally,
the actual outcome $\xs[t] \in \X$ is observed, and everybody suffers
loss as specified by a fixed loss function $\loss \colon \actions \times
\X \to \intcc{0,\infty}$. Thus, the performance of a sequence of actions
$\as[1] \cdots \as[T]$ upon data $\xs[1] \cdots \xs[T]$ is measured
by the cumulative loss $\sum_{t=1}^T \loss(\as[t],\xs[t])$.

\paragraph{Log Loss}

For \emph{log loss} the actions $\actions$ are probability distributions
on $\X$ and $\loss(\pmass,x) = -\log \pmass(x)$, where $\log$ denotes
the natural logarithm. It is important to notice that minimizing log
loss is equivalent to maximizing the predicted probability of outcome
$x$. We write $\ExPred{\xi}_t$ for the prediction of expert $\xi$ at
time $t$ and denote these predictions jointly by $\ExPreds_t$. 

\paragraph{Subsequences} For $m\le n$, we abbreviate $\set{m, \ldots,
n}$ to $m:n$. For completeness, we set $m:n = \emptyset$ for $m > n$.
For any sequence $y_1, y_2, \ldots$ and any set of indices 
$\Cell =
\set{i_1, i_2, \ldots}$ we write $y_\Cell$ for the subsequence
$\family{y_i}{i \in \Cell}$. 
For example, $\xs[\Cell] = \family{x_i}{i \in
\Cell}$ and $\ExPreds_{1:T} = \ExPreds_1, \ldots, \ExPreds_T$. 
If members of a family $\Part = \{\Cell_1, \Cell_2, \ldots\}$ are
pairwise disjoint and together cover $1:T$ ($\Union \Part = 1:T$), then we call
$\Part$ a \emph{partition} of $1:T$, and its members \emph{cells}.

\section{Mixing Past Posteriors}\label{sec:mpp.intro}

Mixing past posteriors (MPP) is a strategy for prediction with expert advice. It operates by maintaining a table of so-called posterior distributions on the set of experts. Each round, we first compute the predictive distribution on experts  by mixing all the posteriors in the table. Then the next outcome is predicted by mixing the expert predictions according to this distribution. Finally, the next outcome is observed. The predictive distribution on experts is conditioned on this outcome, and the posterior distribution thus obtained is appended to the table of posteriors. Note the recursive construction of the distributions in the table; they are not Bayesian posteriors, but conditioned mixtures of all earlier distributions from that same table.

We will not formally introduce MPP here, but recover it as a special case of both the freezing and sleeping algorithms in \secref{sec:mpp.recovered}. Here we state the classical loss bound~\cite[Theorem 7]{bousquet2002}, introducing our notation along the way.  This loss bound relates the loss of MPP to Freund's scheme, where we choose a partition of the data (step~\ref{it:partition.data}) and select an expert for each partition cell (step~\ref{it:choose.experts}). We measure expert performance (step~\ref{it:sum.expert.loss}) using the predictions issued in the context of all data, i.e.\ the traditional interpretation of Freund's scheme.

\subsection{Loss Bound}\label{sec:def.mixing.scheme}
We bound the overhead of MPP over Freund's scheme in terms of the complexity of the reference partition.  We first state the theorem, and then explain the ingredients. We write $\mppprob[w](\xs)$ for the probability that MPP assigns to data $\xs$ (so $- \log \del[\big]{\mppprob[w](\xs)}$ is MPP's cumulative log loss).
\begin{theorem}[{\cite[Theorem 7]{bousquet2002}}]\label{thm:bnw.lossbound} For any mixing scheme $\BackRef$, Bayesian joint distribution $\bayesprob$ with prior distribution $w$ on experts, partition $\Part$ of $1:T$, data $\xs$ and expert predictions $\ExPreds_{1:T}$
\begin{equation}\label{eq:mpp.loss.bound}
\mppprob[w](\xs) ~\ge~  \BackRef(\Part) \bayesprob_\Part(\xs).
\end{equation}
\end{theorem}
\noindent
A \markdef{mixing scheme} $\BackRef$ is a sequence $\BackRef_1, \BackRef_2, \ldots$ of distributions, where $\BackRef_{j+1}$ is a probability distribution on $0:j$. In~\cite{bousquet2002} several mixing schemes are listed, e.g.\ \emph{Uniform Past} and \emph{Decaying Past}. A mixing scheme is turned into a distribution on partitions as follows. Let $\Part$ be a partition of $1:T$, and let $i \in 1:T$.
\markdef{The cell of $i$}, denoted $\Part(i)$, is the unique $\Cell \in \Part$ such that $i \in \Cell$. We write $\prev^\Part(i)$ for \markdef{the predecessor of $i$}, defined as the largest element in $\Part(i) \cup \set{0}$ that is smaller than $i$. Using this notation, the distribution on partitions is given by
\[
\BackRef(\Part) \df \prod_{t \in 1:T} \BackRef_t(\prev^\Part(t)).
\]
Note that this distribution is potentially \emph{defective}; two elements $i < j$ cannot share the same nonzero predecessor, but $\BackRef_i$ may assign nonzero probability to $\prev^\Part(j)$ nonetheless.

Now that we have seen how the loss bound encodes partition, we turn to $\bayesprob_\Part(\xs)$, the probability of the data $\xs$ given a particular partition $\Part$. To compute it, we treat the cells independently \eqref{eq:cells.independent}, and per cell we use the Bayesian mixture with prior $w$ on experts \eqref{eq:bayes.per.cell}, thus mixing the predictions the experts issued in the context of all data \eqref{eq:expert.in.context.of.entire.data}. 
\begin{align}
\label{eq:cells.independent}
\bayesprob_\Part(\xs) &\df \prod_{\Cell \in \Part} \bayesprob_\Part(\xs[\Cell]) \text{, where}
\\
\label{eq:bayes.per.cell}
\bayesprob_\Part(\xs[\Cell]) & \df \sum_{\xi \in \Xi} w(\xi) \ExPred{\xi}_\Cell(\xs[\Cell]) \text{ and}
\\
\label{eq:expert.in.context.of.entire.data}
\ExPred{\xi}_\Cell(\xs[\Cell]) & \df \prod_{i \in \Cell} \ExPred{\xi}_i(\xs[i])
.
\end{align}
A second bounding step allows us to relate the performance of MPP directly to Freund's scheme.  
Let $w$ be the uniform prior over a finite set of experts $\Xi$, and select an expert $\xi^\Cell$ for each partition cell $\Cell \in \Part$.  Then bound each sum \eqref{eq:bayes.per.cell} from below by one of its terms to obtain
\begin{corollary}\label{corl:BnW.two.part.bound} %
\quad \( \displaystyle
\mppprob[w](\xs) 
~\ge~ \BackRef(\Part) 
~\card{\Xi}^{-\card{\Part}}
\prod_{\Cell \in \Part} \ExPred{\xi^\Cell}_\Cell(\xs[\Cell])
.
\)
\end{corollary}

\noindent
Thus the log-loss overhead of MPP over Freund's scheme is bounded by $- \log \BackRef(\Part) + \card{\Part} \log \card{\Xi}$, which can be related to the number of bits to encode the chosen partition and the selected experts for each cell~\cite{bousquet2002}.

\paragraph{Convex Combinations}
In~\cite{bousquet2002}, the authors make a point of selecting a
\emph{convex combination of experts} for each subsequence, where the
loss of a convex combination of experts is the weighted average
\emph{loss} of the experts. The loss of such a convex combination is therefore \emph{always} higher than the loss of its best expert. Uniform bounds in terms of arbitrary experts, like \corlref{corl:BnW.two.part.bound}, apply in particular to the best expert, and hence to any convex combination. Therefore, without loss of generality, we do not discuss convex combinations any further.

\paragraph{Interpreting Freund's Problem}
This loss bound shows that MPP solves the black-box-experts interpretation of Freund's problem. This can be seen clearly in \eqref{eq:expert.in.context.of.entire.data}. To predict the subsequence $\xs[\Cell]$, it uses predictions $\ExPred{\xi}_\Cell$ which were issued in the context of all data. This means that the experts observe the entire history $\xs[1:i]$ before predicting the next outcome $\Xs[i+1]$. 

Switching between \emph{learning} experts that observe all data is useful when the data are homogeneous, and the experts learn its global pattern at different speeds. In such cases we want to train each expert on all observations, for then by switching at the right time, we can predict each outcome using the expert that has learned most \emph{until then}. This scenario is analysed in~\cite{threemusketeers07}, where experts are parameter estimators for a series of statistical models of increasing complexity.

On the other hand, if the data have local patterns then our new interpretation of Freund's problem applies, and we want to train each expert on the subsequence on which it is evaluated, so that it can 
exploit its local patterns. To solve Freund's problem for such learning experts, we need to know about its internal structure.

\section{Structured Experts}\label{sec:ehmm}

\exclude{
TODO: Refer to \cite{gyorgy2005}; For sleeping we are doing something
similar.
}

Assume there is only a single expert and fix a reference partition.
Suppose we want to predict as if the expert is restarted on each cell
of the partition, when in reality the expert just makes her predictions
as if all the data were in a single cell. Then clearly this is impossible
if we treat this expert completely as a black box: if we do not know
what the expert's predictions would have been if a certain outcome were,
say, the start of a new cell, then we cannot match these predictions.

The expert therefore needs to reveal to us some of her internal state.
To this end, we will represent the parts of her internal state that will
\emph{not} be revealed to us by lower level experts that we will treat
as black boxes, and assume our main expert combines the predictions of
these base experts using an \emph{expert hidden Markov model} (EHMM).

\subsection{EHMMs}

\begin{figure}
\caption{Bayesian Network specification of an EHMM\label{fig:ehmm}}
\centering
\footnotesize
$\xymatrix@!0@R=3.0em@C=3.5em{
\lt{\ar[r]}{\pinit} &
\rv{\Qs[1]} \lt{\ar[r]}{\ptf} \lt{\ar[d]}{\ppf} & 
\rv{\Qs[2]} \lt{\ar[r]}{\ptf} \lt{\ar[d]}{\ppf} &
\rv{\Qs[3]} \lt{\ar[r]}{\ptf} \lt{\ar[d]}{\ppf} & 
\rv{\Qs[4]} \lt{\ar[r]}{\ptf} \lt{\ar[d]}{\ppf} & \cdots
\\
&
\rv{\Xis[1]} \lt{\ar[d]}{\ExPreds_1} & 
\rv{\Xis[2]} \lt{\ar[d]}{\ExPreds_2} & 
\rv{\Xis[3]} \lt{\ar[d]}{\ExPreds_3} &
\rv{\Xis[4]} \lt{\ar[d]}{\ExPreds_4} & \cdots
\\
&
\rv{\Xs[1]} & 
\rv{\Xs[2]} & 
\rv{\Xs[3]} &
\rv{\Xs[4]} & \cdots
}
$
\end{figure}
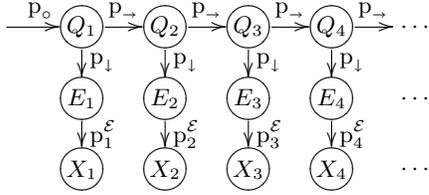

Expert Hidden Markov Models (EHMMs) were introduced in~\cite{koolen08:_combin_exper_advic_effic} as a language to specify
strategies for prediction with expert advice. We briefly review them
here. An \markdef{EHMM} $\A$ is a probability distribution that is
constructed according to the Bayesian network in \figref{fig:ehmm}. 
It is used to sequentially predict outcomes $\Xs[1]$, $\Xs[2]$,
$\ldots$ which take values in outcome space $\X$. At each time $t$, the
distribution of $\Xs[t]$ depends on a hidden state $\Qs[t]$, which
determines mixing weights for the experts' predictions. Formally, the
\markdef{production function} $\ppf$ determines the interpretation of a
state: it maps any state $\qs[t] \in \Q$ to a distribution $\ppf[{\qs[t]}]$ on the identity $\Expert_t$ of the expert that should be
used to predict $\Xs[t]$. Then given $\Expert_t = \expert$, the
distribution of $\Xs[t]$ is base expert $\expert$'s prediction
$\ExPred{\expert}_t$. It remains to define the distribution of the hidden
states. The starting state $\Qs[1]$ has \markdef{initial distribution}
$\pinit$, and the state evolves according to the \markdef{transition
function} $\ptf$, which maps any state $\qs[t]$ to a distribution
$\ptf[{\qs[t]}]$ on states. 

An EHMM $\A$ defines a prediction strategy as follows; after observing
$\xs[1:t]$, predict outcome $\Xs[t+1]$ using the marginal
$\A\delc{\Xs[t+1]}{\xs[1:t]}$, which is a \emph{mixture} of the expert's
predictions $\ExPreds_{t+1}$.

\exclude{

It may seem like a hindrance to restrict the set of states $\Q$ to be
the same at each time. However, one may perfectly well construct a
separate set of states for each time step, and include all these sets
in $\Q$. By setting up $\ptf$ to only allow transitions from states
associated with time $t$ to states associated with time $t+1$, the
states effectively all get their own time. See~\cite{DeRooijVanErven2009,koolen08:_combin_exper_advic_effic} for
examples.
}

\newcommand{\slotstate}[1]{\textsf{#1}}
\begin{figure}
  \caption{Hidden state transitions in slot machine HMM\label{fig:slotmachine}}
  \centering
  \footnotesize
  $\xymatrix@!0@R=3.0em@C=2.5em{%
    & \slotstate{Jackpot} \ar[dl]\\
    \slotstate{Cold}\ar@(dl,ul)^{\nicefrac{99}{100}} \ar[rr]_{\nicefrac{1}{100}} && \slotstate{Hot} \ar@(dr,ur)_{\nicefrac{9}{10}}[]
    \ar[ul]_{\nicefrac{1}{10}}
  }$
\end{figure}
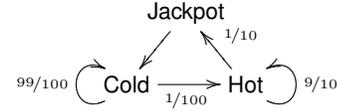

\begin{example}[Any Ordinary HMM]
  To illustrate how ordinary HMMs are a special case of EHMMs, consider
  the following naive gambler's HMM model of an old-fashioned slot
  machine: in each round the gambler inserts one nickel into the slot
  machine and then the machine pays out a certain number of nickels
  depending on its hidden internal state: in state \slotstate{Cold} it
  pays out nothing; in state \slotstate{Hot} it pays out an amount
  between one and five nickels, uniformly at random; and then there's
  \slotstate{Jackpot} in which it always pays out ten nickels. The
  machine always starts in state \slotstate{Cold} and the state
  transitions are as in \figref{fig:slotmachine}.

  To make an EHMM out of this HMM, we just identify experts with states:
  $\Q = \Xi = \{\slotstate{Cold},\slotstate{Hot},\slotstate{Jackpot}\}$,
  $\ppf[\e](\e) = 1$, and each expert predicts according to the
  corresponding payout scheme.
  The distributions on states follow the original HMM:
  $\pinit(\slotstate{Cold})=1$ and $\ptf$ as in
  \figref{fig:slotmachine}.
\end{example}

\begin{example}[Bayes on base experts]\label{ex:bayes.on.experts}
  We identify the \\Bayesian distribution with prior $w$ on base experts
  $\Xi$ and the EHMM with $\Q = \Xi$, $\pinit = w$, and $\ptf[\e](\e) =
  \ppf[\e](\e) = 1$, since their marginals coincide. Despite its
  deceptive simplicity, this EHMM \emph{learns}: its marginal
  distribution on the next outcome is a mixture of the expert's
  predictions according to the Bayesian posterior.
\end{example}

\begin{example}[Bayes on EHMMs]
  Fix EHMMs $\A^1, \ldots, \A^n$ with disjoint state spaces and the
  same basic experts, and let $w$ be a prior distribution on $1:n$. The
  Bayesian mixture EHMM has state space $\Q = \bigcup_{i} \Q^i$, and for
  any two states $\q,\q' \in \Q^i$ belonging to the same original EHMM,
  $\pinit(\q) = w(i) \pinit^i(\q)$, $\ptf[\q](\q') = \ptf[i,\q](\q')$
  and $\ppf[\q](\e) = \ppf[i,\q](\e)$. Again, this EHMM \emph{learns}
  which of the given EHMMs is the best predictor.
\end{example}

\subsection{The Forward Algorithm}
Sequential predictions for EHMMs can be computed efficiently using the
\emph{forward algorithm}, which maintains a posterior distribution over
states, and predicts each outcome with a mixture of the experts'
predictions~\cite{koolen08:_combin_exper_advic_effic}. Given a posterior
$\PredPost_t(\Qs[t]) = \A(\Qs[t] | \xs[1:t-1])$ for the hidden state at
time $t$, the forward algorithm predicts $\xs[t]$ using the marginal of
$\A(\Qs[t],\Es[t],\Xs[t] | \xs[1:t-1])$. Then, after observing outcome
$\xs[t]$, it updates its posterior $\PredPost_t$ for $\Qs[t]$ to a
posterior $\PredPost_{t+1}$ for $\Qs[t+1]$.

For finite $\Q$, $\expertset$ and $\X$, the running time of the
algorithm is determined by this last posterior update step, which in
general may require $O\del[\big]{|\Q|^2}$ computation steps for each
round $t$. On $T$ outcomes, this gives a total running time of
$O\del[\big]{|\Q|^2\cdot T}$. In \appref{app:runningtime} we provide a
more careful analysis.

\newcommand{\B}{\hmm{B}}
\newcommand{\C}{\hmm{C}}

\section{Freezing \& Sleeping}\label{sec:freezingandsleeping}
Let $\xs = \xs[1], \ldots, \xs[T]$ be a sequence of data and suppose
that a reference partition $\Part$ of $1:T$ is given in advance. We are
interested in the performance of a structured expert $\A_\Part$, which
for each cell $\Cell \in \Part$
runs a separate instance of the structured expert $\A$ on the subsequence
$\xs[\Cell]$. This leaves unspecified,
however, whether the original timing of $\xs[\Cell]$ should be preserved
when $\xs[\Cell]$ is presented to $\A$. This is a modelling choice,
which depends on the application at hand. We therefore treat both the
case where the timing is preserved, which we call \emph{sleeping}, and
the case where the timing is not preserved, which we call
\emph{freezing}. (See also \figref{fig:mot.ex} in the introduction.)

\paragraph{Sleeping}

\begin{figure}
\caption{Sleeping and Freezing on outcomes $\xs[\{2,4,5,\ldots\}]$}
\subfloat[Sleeping: EHMM $\A^\sl_{\{2,4,5,\ldots\}}$\label{fig:sleeping}]{
\footnotesize
$\xymatrix@!0@R=3.0em@C=3.5em{
\lt{\ar[r]}{\pinit} &
\rv{\Qs[1]} \lt{\ar[r]}{\ptf}  & 
\rv{\Qs[2]} \lt{\ar[r]}{\ptf} \lt{\ar[d]}{\ppf} & 
\rv{\Qs[3]} \lt{\ar[r]}{\ptf}  & 
\rv{\Qs[4]} \lt{\ar[r]}{\ptf} \lt{\ar[d]}{\ppf} & 
\rv{\Qs[5]} \lt{\ar[r]}{\ptf} \lt{\ar[d]}{\ppf} & \cdots
\\
&
& 
\rv{\Xis[2]} \lt{\ar[d]}{\ExPreds_2} & 
& 
\rv{\Xis[4]} \lt{\ar[d]}{\ExPreds_4} & 
\rv{\Xis[5]} \lt{\ar[d]}{\ExPreds_5} & \cdots
\\
&
& 
\rv{\Xs[2]} & 
& 
\rv{\Xs[4]} &
\rv{\Xs[5]} & \cdots
}$}%
\\%
\subfloat[Freezing: EHMM $\A^\fr_{\{2,4,5,\ldots\}}$\label{fig:freezing}]{
\footnotesize
$\xymatrix@!0@R=3.0em@C=3.5em{
\lt{\ar[rr]}{\pinit}&
& 
\rv{\Qs[2]} \lt{\ar[rr]}{\ptf} \lt{\ar[d]}{\ppf} & 
& 
\rv{\Qs[4]} \lt{\ar[r]}{\ptf} \lt{\ar[d]}{\ppf} &
\rv{\Qs[5]} \lt{\ar[r]}{\ptf} \lt{\ar[d]}{\ppf} & \cdots
\\
&
& 
\rv{\Xis[2]} \lt{\ar[d]}{\ExPreds_2} & 
& 
\rv{\Xis[4]} \lt{\ar[d]}{\ExPreds_4} & 
\rv{\Xis[5]} \lt{\ar[d]}{\ExPreds_5} & \cdots
\\
&
& 
\rv{\Xs[2]} & 
& 
\rv{\Xs[4]} & 
\rv{\Xs[5]} & \cdots
}$}%
\end{figure}

We say that the instance of $\A$ that is used to predict cell $\Cell$ is sleeping if it does
notice the passing of time during outcomes outside of $\Cell$, even
though it does not observe them. We write $\sleepcomp$ for the resulting
EHMM, which is shown in \figref{fig:sleeping} for the example $\Cell =
\set{2,4,5,\ldots}$. Notice that $\sleepcomp$ contains all five states
$\Qs[1:5]$, even though it does not observe $\xs[1]$ or $\xs[3]$. This
has the effect that state transitions from e.g.\ $\Qs[2]$ to $\Qs[4]$
are composed of two transition steps according to $\ptf$. The
distributions on individual cells combine into the following
distribution on all data $\xs$:
\begin{equation*}
  \sleepcomp[\Part](\xs) \df
    \prod_{\Cell \in \Part} \sleepcomp(\xs[\Cell]).
\end{equation*}
To memorize the nature of sleeping, one may think of the way television
channels get interleaved as you zap between them: a channel not being
watched is not paused, but instead continues broadcasting even when its
content is not observed.

\paragraph{Freezing}

In freezing, the instance of $\A$ that is used to predict cell $\Cell \in \Part$ is frozen
when outcomes outside of $\Cell$ occur: its internal state should not
change based on those outcomes. (Of course we have no control over the
base experts on which $\A$ is based, so they may do whatever they please
with such data. We therefore do have to preserve the timing of the base
experts' predictions.) The resulting EHMM $\freezcomp$ is shown for the
example $\Cell = \set{2,4,5,\ldots}$ in \figref{fig:freezing}. Note that
$\Qs[2]$, $\Qs[4]$ and $\Qs[5]$ are the first, second and third state of
$\freezcomp$; state transitions between them consist of a single
transition step according to $\ptf$. The resulting distribution on all
data is defined by
\begin{equation*}
  \freezcomp[\Part](\xs) \df
    \prod_{\Cell \in \Part} \freezcomp(\xs[\Cell]).
\end{equation*}
One might associate freezing with the way different e-mail conversations
get interleaved in your inbox (if it is sorted by order of message
arrival): a conversation about your latest research is paused (remains
frozen) regardless of how much spam you receive in between.

\subsection{An Infeasible Solution}

The freezing or sleeping distributions can be computed if the reference
partition $\Part$ is given in advance. The problem we are addressing,
however, is that we do not assume $\Part$ to be known. An easy (but
impractical) solution to this problem is to predict according to the
Bayesian mixture of all possible partitions: let $w$ be a prior on the
set of all possible partitions and predict such that the joint
distribution on all data is given by
\begin{equation*}
  \B(x) \df \sum_{\Part} w(\Part) \A^\fos_\Part(x),
\end{equation*}
where $\fos$ denotes either $\fr$ for freezing or $\sl$ for sleeping. Lower bounding the sum by the
term for the reference partition $\Part$ directly gives an upper bound
on the log loss:
\begin{equation*}
  -\log \B(x) \leq -\log w(\Part) -\log \A^\fos_\Part(x).
\end{equation*}
To predict according to $\B$ in general would require an exponential
amount of state to keep track of all possible partitions, which is
completely impractical. In the following section we therefore present
generalisations to both sleeping and freezing of the mixing past
posteriors algorithm and show that their running time is comparable to that of the forward algorithm on $\A$ itself. Then in section
\secref{sec:lossbounds} we prove bounds that relate the additional loss
to the encoding cost of the reference partition $\Part$.

\subsection{The \OurAlgNotFancy{} Algorithm}\label{sec:the.algorithm}

\begin{algorithm}[tb]
\caption[\OurAlg{}]{\OurAlg{}: Evolving Past Posteriors\label{alg:main}}
\smallskip
 \textbf{Input:}
\begin{itemize} \setlength{\itemsep}{0pt}
\item An EHMM $\A$ with components $\pinit$, $\ptf$ and $\ppf$ (see \secref{sec:ehmm}) 
\item A mixing scheme $\BackRef_1, \BackRef_2, \ldots$ (see \secref{sec:def.mixing.scheme} and \secref{sec:mixing.schemes})
\item Expert predictions $\ExPreds_1, \ExPreds_2, \ldots$ and data $x_1, x_2, \ldots$
\end{itemize}
\textbf{Output:}
Predictions $\MyPred_1, \MyPred_2, \ldots$\\
\textbf{Storage:} %
Past posteriors $\PastPost_1, \PastPost_2, \ldots$ on $\Q$, the states of $\A$
\rule{\columnwidth}{.25pt}

\smallskip
\textbf{Algorithm}\\
{\renewcommand{\arraystretch}{1.5}
\newcounter{linenoctr}
\setcounter{linenoctr}{0}
\newcommand{\lineno}{\refstepcounter{linenoctr}\arabic{linenoctr}}
\newcommand{\mathline}[2]{\multicolumn{2}{r}{$\displaystyle#1$} &$\gets$ & $\displaystyle#2$}
\newcommand{\mathlinei}[2]{$\displaystyle#1$ &$\gets$ & $\displaystyle#2$}
\newcommand{\textlinei}[1]{\multicolumn{3}{l}{\parbox[t]{.85\columnwidth}{#1}}}
\begin{tabular}{l@{}lr@{~}c@{~}l}
\lineno:
& \multicolumn{4}{l}{\parbox[t]{.9\columnwidth}{Set the first posterior to the initial distribution of $\A$}}
\\
& \mathline{\PastPost_1(\qs[1])}{\pinit(\qs[1])}
\\
\lineno: 
& \multicolumn{4}{l}{\textbf{for} $t=1,2,\ldots$ \textbf{do}}
\\
\lineno: \label{line:lambda}
&& \textlinei{Form $\PredPost_t$, the current configuration, as the $\BackRef_t$-mix\-ture of past posteriors:}
\\
&&\mathlinei{\PredPost_t(\qs[t])}{\sum_{0 \le j < t} \BackRef_t(j) \PastPost_{j+1}(\qs[t]).}
\\
\lineno:
&& \textlinei{Compute $\Joint$, the joint distribution on states, experts and outcomes:}
\\
&& \mathlinei{\Joint(\qs[t],\es[t],x_t)}{\PredPost_t(\qs[t])  \ppf[{\qs[t]}](\e_t) \ExPred{\e_t}_t(x_t).}
\\
\lineno:
&& \textlinei{Predict $x_t$ using the marginal $\Joint(x_t)$,}
\\
\lineno:
&& \textlinei{Observe $x_t$. Suffer log loss}
\\
&& \mathlinei{\loss^\Alg_t}{- \log \del[\big]{\MyPred_t(x_t)}.}
\\
\lineno: \label{line:loss.update}
&& \textlinei{Perform loss update and state evolution to obtain the next posterior}
\\
&& \mathlinei{\qquad \PastPost_{t+1}(\qs[t+1])}{\sum_{\qs[t] \in \Q} \Joint(\qs[t]|x_t) \ptf[{\qs[t]}](\qs[t+1]).}
\\
\lineno: \label{line:only.for.sleeping}
&& \textlinei{Only for sleeping: perform state evolution for all past posteriors $(1 \leq j \leq t)$}
\\
&& \mathlinei{\PastPost_j(\qs[t+1])}{\sum_{\qs[t] \in \Q} \PastPost_j(\qs[t])
\ptf[{\qs[t]}](\qs[t+1]).}
\\
\lineno: & \multicolumn{4}{l}{\textbf{end for}}
\end{tabular}
}
\end{algorithm}

Here we present a generalisation of the mixing past posteriors (MPP)
algorithm, which we call \emph{evolving past posteriors} (\OurAlg{}). It
is based on the view that MPP internally uses the Bayesian mixture of
base experts, which is a standard EHMM. Given this perspective and after
making the distinction between sleeping and freezing, the generalisation
to other EHMMs is straightforward. We will discuss the connections
between MPP and \OurAlg{} in more detail in \secref{sec:mpp.recovered}.

The \OurAlg{} algorithm has variants for sleeping and freezing, which are both
given in \algoref{alg:main}. It takes an EHMM $\A$ and mixing scheme
$\BackRef$ (see \secref{sec:def.mixing.scheme}) as input. Given a
distribution $\PredPost_t$ on the hidden state $\Qs[t]$ at time $t$, the
\OurAlg{} algorithm predicts $\Xs[t]$ exactly like the forward
algorithm. It differs from the forward algorithm, however, in the way it
computes $\PredPost_t$. Whereas in the forward algorithm $\PredPost_t$
may be interpreted as the posterior distribution on $\Qs[t]$, in the
\OurAlg{} algorithm $\PredPost_t$ is a $\BackRef$-mixture of \emph{the
algorithm's own past posteriors}. This recursive nature of \OurAlg{},
which it inherits from the MPP algorithm, makes it hard to analyse.

We denote by $\frprob[\A]$ and $\slprob[\A]$ the probability distributions on $\family{\Qs[t], \Es[t], \Xs[t]}{t \in \N}$ defined by \FreezAlg{} and \SleepAlg{} on EHMM\/ $\A$ and mixing scheme $\BackRef$. For both $\fos \in \set{\sl,\fr}$
\[
\fosprob[\A](\qs,\es,\xs) ~=~ \prod_{t \in 1:T} \Joint(\q_t,\e_t,\x_t).
\]

\subsubsection{Representation Invariance}
Let $\A^1$ and $\A^2$ be EHMMs that are based on the same set of experts $\Xi$, but have different state spaces. We call $\A^1$ and $\A^2$ \markdef{equivalent} if $\A^1(\es) = \A^2(\es)$ for all $\es$. Consequently, equivalent EHMMs assign the same probability $\A^1(\xs) = \A^2(\xs)$ to all data $\xs$, hence the difference between $\A^1$ and $\A^2$ is merely a matter of \emph{representation}. As an important sanity check, we need to verify that \OurAlg{} on either EHMM issues the same predictions. 

\begin{theorem}[Invariance]\label{thm:invariance}
Let \fos{} denote either \fr{} or \sl{}. Fix equivalent EHMMs $\A^1$ and $\A^2$. Then for all data $\xs$
\[
\fosprob[\A^1](\xs) ~=~ \fosprob[\A^2](\xs).
\]
\end{theorem}
\begin{proof} Given in \appref{app:invariance}.
\end{proof}
\noindent
Thus, from the perspective of predictive performance, the difference between $\A^1$ and $\A^2$ is irrelevant. 
Of course, it does matter for the computational cost of \OurAlg{}, see \secref{sec:running.time}.

\subsubsection{Mixing Schemes}\label{sec:mixing.schemes}
\begin{table}
\caption{Mixing schemes}\label{tab:mixing.schemes}
\begin{tabular}{llll}
Mixing scheme  & $\BackRef_{t+1}(t)$ & $\BackRef_{t+1}(j)$ for $0 \le j < t$
\\
\midrule
Yesterday & $1$ & $0$
\\
Fixed Share$(\alpha)$ & $1-\alpha$ & $\alpha$ if $j=0$ and $0$ o.w.
\\
Uniform past$(\alpha)$ & $1-\alpha$ & $\alpha/t$
\\
Decaying past$(\alpha,\gamma)$& $1-\alpha$ & $\alpha(t-j)^{-\gamma}/Z_t$
\end{tabular}
\end{table}

Bousquet and Warmuth~\cite{bousquet2002} provide an extensive discussion
of possible mixing schemes. Their loss bounds for various schemes carry
over directly to our setting. It is interesting, however, to analyse the
running times of the Fixed-Share to \emph{uniform past} and to
\emph{decaying past} mixing schemes for \OurAlg{}\@. For further
information we refer the reader to~\cite{bousquet2002}.

Both schemes (see \tabref{tab:mixing.schemes}) depend on a \emph{switching rate} $\alpha \in [0,1]$, which
determines whether to continue with yesterday's posterior or switch back
to an earlier one: $\BackRef_{t+1}(t) = 1 - \alpha$ and $\sum_{0 \leq j
< t} \BackRef_{t+1}(j) = \alpha$.

\paragraph{Uniform Past} Given the choice to switch back, the uniform
past mixing scheme gives equal weights to the entire past:
$\BackRef_{t+1}(j) = \alpha/t$ for $0 \leq j < t$.

\paragraph{Decaying Past} Instead, the decaying past scheme assigns larger
weight to the recent past: $\BackRef_{t+1}(j) = \alpha
(t-j)^{-\gamma}/Z_t$ for $0 \leq j < t$, where $Z_t = \sum_{0 \leq j <
t} (t-j)^{-\gamma}$ is a normalising constant and $\gamma \geq 0$ is a
parameter that determines the rate of decay.

\subsubsection{Running Times}\label{sec:running.time}

\appref{app:runningtime} provides a detailed comparison of the running
times and space requirements of \OurAlg{} and the forward algorithm. The
upshot is that for the uniform past mixing scheme the sleeping variant
of \OurAlg{} is as efficient as the forward algorithm, in terms of both
running time and space requirements; the freezing variant is equally
efficient if the set of hidden states $\Q$ is finite, but may be a
factor $O(T)$ less efficient on $T$ outcomes for countably infinite
$\Q$. The decaying past mixing scheme is a factor $O(T)$ less efficient
(for both time and space) than uniform past in all cases, but may be
approximated by a scheme described in~\cite{bousquet2002} that reduces
this factor to $O(\log T)$.

\subsection{Loss Bound}\label{sec:lossbounds}

This bounds relates the performance of \FreezAlg{} and \SleepAlg{} (defined in \algoref{alg:main}) to that of $\freezcomp[\Part]$ and $\sleepcomp[\Part]$ for all partitions $\Part$ \emph{jointly}.

\begin{theorem}[\OurAlg{} Loss Bounds]\label{thm:loss.bound}
  For both $\fos \in \set{\fr,\sl}$ and any mixing scheme $\BackRef$, data $\xs$ and expert predictions $\ExPreds_{1:T}$
\begin{align}
  \fosprob[\A](\xs[1:T])
    &~\ge~  \sum_{\Part} \BackRef(\Part)
      \foscomp[\Part](\xs[1:T]) \label{eqn:fosbound}.
\end{align}
\end{theorem}
\begin{proof}
Given in \appref{app:loss.bound.proof}.
\end{proof}
\noindent
Using this bound, we can relate the predictive performance of \SleepAlg{} and \FreezAlg{} to that of $\sleepcomp[\Part]$ and $\freezcomp[\Part]$ for any reference partition $\Part$.
\begin{corollary}\label{crl:twopart.loss.bound}
\quad $
  \fosprob[\A](\xs[1:T])
    ~\ge~ \BackRef(\Part)  \foscomp[\Part](\xs[1:T])
$.
\end{corollary}
\noindent
From the brutal way in which \corlref{crl:twopart.loss.bound} was obtained, we may expect to often do much better in practice; \emph{many} partitions may contribute significantly to \eqref{eqn:fosbound}. %

\subsection{Recovering MPP}\label{sec:mpp.recovered}
We now substantiate our claim that \OurAlg{} generalises MPP by proving that MPP results from running \FreezAlg{} or \SleepAlg{} on the Bayesian EHMM (\exref{ex:bayes.on.experts}).

\begin{theorem}\label{thm:mpp.recovered}
Let $\A$ be the Bayesian EHMM with initial distribution $w$, and let $\mppprob[w]$ denote the probability distribution defined by MPP with prior $w$. Then for all data $\xs$
\[
\frprob[\A](\xs) ~=~ \slprob[\A](\xs) ~=~ \mppprob[w](\xs).
\]
\end{theorem}

\begin{proof}
The difference between freezing and sleeping (line~\ref{line:only.for.sleeping}) evaporates since state evolution is the identity operation. By identifying states and experts the MPP algorithm~\cite[Figure 1]{bousquet2002} remains.
\end{proof}

\noindent
The theorem does not require the set of experts $\E$ to be finite. If $\E$ is infinite (or too large), MPP is intractable. Still, a small EHMM may exist that implements Bayes (say with the uniform prior) on $\E$, and we can use \SleepAlg{} (which is faster than \FreezAlg{}) for sequential prediction. For example, we may implement MPP on the infinite set of Bernoulli experts (cf\ the example in the introduction) efficiently, in time $O(T^2)$, using \SleepAlg{} on the \emph{universal element-wise mixture} EHMM of~\cite[\S 4.1]{koolen08:_combin_exper_advic_effic}.

\subsubsection{Improved MPP Loss Bound}
\cite[Theorem 7]{bousquet2002}~(our \thmref{thm:bnw.lossbound}) bounds the overhead of MPP over Freund's scheme in terms of $\BackRef(\Part)$, the complexity of the reference partition $\Part$ according to the mixing scheme $\BackRef$. A more general bound follows directly from Theorems~\ref{thm:loss.bound} and~\ref{thm:mpp.recovered}:

\begin{corollary}\label{corl:mpp.one.part}
\quad
$\displaystyle
\mppprob[w](\xs) ~\ge~  \sum_\Part \BackRef(\Part) \bayesprob_\Part(\xs)
$.
\end{corollary}
\noindent
Even with a fixed reference partition $\Part$ in mind, we get a better bound by considering small modifications of $\Part$, e.g.\ finer partitions or partitions that disagree about a single round.

\paragraph{Adversarial Experts}
For each number of rounds $T$ one can construct a set of $T$ base experts and data $\xs$ such that the loss of Freund's scheme under the MPP interpretation is infinite for all partitions except the finest one. We simply have expert $t$ suffer infinite loss in all rounds other than $t$. In this pathological case the bounds in \thmref{thm:bnw.lossbound} for that partition and \corlref{corl:mpp.one.part} are equal and tight.%

\subsubsection{Is \OurAlgNotFancy{} strictly more general than MPP?}
A natural question is whether either \SleepAlg{} or \FreezAlg{} can be implemented using MPP on a rich set of meta-experts. To preclude the trivial answer that regards either algorithm as a single meta-expert, we ask for a fixed construction that works for all mixing schemes.

\paragraph{Sleeping}
For any EHMM $\A$, \SleepAlg{} can be reduced to MPP on meta-experts.
Let the set of meta-experts be $\Q^\infty$, the set of paths through the
hidden states of $\A$. Each meta-expert $\q_\N$ predicts $\xs[t]$ using
the $\ppf[{\qs[t]}]$-mixture of base expert predictions. We set
the prior $w$ in MPP equal to the marginal probability measure of $\A$
on paths (as determined by $\pinit$ and $\ptf$). We omit the proof that
the predictions made by MPP on these meta-experts with prior $w$ are
equal to those made by \OurAlg{} on $\A$.

\paragraph{Freezing}
\newcommand{\ea}{\textnormal{a}}
\newcommand{\eb}{\textnormal{b}}
The next example shows that \FreezAlg{} really is more general than MPP. Fix two experts $\E = \set{\ea{},\eb{}}$. Consider the EHMM $\A$ that predicts the first outcome using expert \ea{}, and the second outcome using expert \eb{}, i.e.\ $\Q=\E$, and $\pinit(\ea{}) = \ptf[\ea{}](\eb{}) = \ppf[\q](\q) = 1$. Running \FreezAlg{} on $\A$ results in $\PastPost_2(\eb{}) = \PastPost_1(\ea{}) = 1$, so that the first outcome is predicted using expert \ea{}, and the second outcome is predicted using the $\BackRef_2$-mixture of experts. Thus any candidate meta-expert \emph{must} predict the first outcome using base expert \ea{}. But that means that for MPP with prior $w$ on meta-experts, the loss update has no effect, so that $w = \PastPost_1 = \PastPost_2 = \PredPost_2$. Hence the second outcome will be predicted according to the prior mixture of experts. Since $\BackRef_2$ is arbitrary and $w$ is fixed, there can be no general scheme to reduce \FreezAlg{} to MPP.

\newcommand{\predfunction}{\operatorname{Pred}}
\newcommand{\palg}{\pmass^\Alg}
\newcommand{\Palg}{\Pmass^\Alg}
\section{Other Loss Functions}\label{sec:mixable.losses}

We will now show how the \OurAlg{} algorithm for logarithmic loss can be
directly translated into an algorithm with corresponding loss bound for
any other mixable loss function. The same construction works for any
logarithmic loss algorithm that predicts according to a mixture of the
experts' predictions at each trial and whose predictions only depend on
the experts' past losses on outcomes that actually occurred.

\paragraph{Mixability} A loss function $\loss \colon \actions \times
\X \to \intcc{0,\infty}$ is called
\emph{$\eta$-mixable} for $\eta > 0$ if any distribution $p$ on
experts $\expertset$ can be mapped to a single action
$\predfunction(p) \in \actions$ in a way that guarantees that
\begin{equation}\label{eqn:mixability}
  \loss\del[\big]{\predfunction(p),x}
    \leq -\tfrac{1}{\eta} \log \ex_{\expert \sim p}
      \Big[\exp\del[\big]{-\eta \loss(\ExAct{\expert}\!, x)}\Big]
\end{equation}
for all outcomes $x \in \X$ and expert predictions $\ExActs$. It is called
\emph{mixable} if it is $\eta$-mixable for some $\eta > 0$~\cite{cesa-bianchi2006}. Mixability ensures that
expert predictions for $\loss$ loss can be mixed 
in essentially the same way as for log loss.

For example, logarithmic loss itself is $1$-mixable. And for $\actions =
[0,1]$ and $\X = \{0,1\}$ the \emph{square loss} $\loss(\a,x) \df
(\a-x)^2$ is $2$-mixable and the \emph{Hellinger loss} $\loss(\a,x) \df$\\$
((\sqrt{1-x} - \sqrt{1-\a})^2 + (\sqrt{x} - \sqrt{\a}))/2$ is
$\sqrt{2}$-mixable.\cite{haussler1998,cesa-bianchi2006}

\paragraph{The Benefits of Lying} 
Given data $\xs[1:t]$ and expert predictions $\ExActs_{1:t}$, 
let $\loss^\expert_{1:t} \df \loss( \a^{\expert}_1, x_1),
\ldots, \loss(\a^{\expert}_t, x_t)$ denote the sequence of losses 
of expert $\expert$, and
let $\loss^\expertset_{1:t}$ denote these losses jointly for all
experts. In the special case that $\loss$ is the logaritmic loss we
write $\logloss^\expert_{1:t}$ and $\logloss^\expertset_{1:t}$,
respectively.

\newcommand{\walg}[3][nope]{%
\ifthenelse{\equal{#1}{nope}}{%
\palg_{#2}\sbr{#3}}{%
\palg_{#2}\sbr{#3}(#1)}%
}
\newcommand{\pastt}{{<t}}
\newcommand{\pastT}{{<T}}
\newcommand{\past}{\xs[\pastt],\logloss^\expertset_\pastt}

Suppose \AlgAlg{} is an algorithm for log loss that predicts each outcome $\xs[t]$ by mixing the experts' predictions $\ExPreds_t$ according to the
distribution $\walg{t}{\past}$ on \emph{experts}. The square-bracket expression indicates that $\palg_t$ may depend on the past outcomes $\xs[1:t-1]$ and the losses of the
experts on these outcomes, but not on the experts' past or current predictions in
any other way. Following this convention, the algorithm predicts $\xs[t]$ using:
\begin{equation*}
\walg[\x_t]{t}{\past} 
~\df~
\sum_\expert \walg[\expert]{t}{\past} \ExPred{\expert}_t(\xs[t]).
\end{equation*}
Now for any game with $\eta$-mixable loss $\loss$ and the same set of
experts $\expertset$, we can derive from \AlgAlg{} an algorithm
$\AlgAlg^\eta_{\loss}$ that predicts $\xs[t]$ according to
\begin{equation*}
\a^{\Alg^\eta_{\loss}}_t 
~\df~
\predfunction\big(\walg{t}{\xs[\pastt], \eta \loss^\expertset_\pastt}\big).
\end{equation*}
Note that $\AlgAlg^\eta_{\loss}$ is lying to \AlgAlg{}: while \AlgAlg{} thinks it
is playing a game for log loss in which experts have incurred log losses
$\eta \loss^\expertset_\pastt$, in reality $\AlgAlg^\eta_{\loss}$ is playing
a game for loss $\loss$ and is feeding \AlgAlg{} fake inputs and
redirecting \AlgAlg{}'s outputs.
Let us now analyse the loss of the derived algorithm
$\AlgAlg^\eta_{\loss}$.

\begin{lemma}[Other Loss Functions]
  \label{lem:otherlosses}
  Suppose \AlgAlg{} is an algorithm for logarithmic loss that predicts
  according to\\$\walg{t}{\past}$ at each time $t$, $\loss$ is an
  $\eta$-mixable loss function, and $f(\xs[1:T],\loss^\expertset_{1:T})$
  is an arbitrary function that maps outcomes and expert losses to real
  numbers. Then any log loss bound for \AlgAlg{} of the form
  \begin{align}\label{eqn:loglossbound}
    -\log \Palg(\xs[1:T])
    &~\leq~ f(\xs[1:T],\logloss^\expertset_{1:T}) 
    &\text{for all $\ExPreds_{1:T}$},
  \intertext{
    directly implies the $\loss$ loss bound for $\AlgAlg^\eta_{\loss}$:
  }
    \loss(\a^{\Alg^\eta_{\loss}}_{1:T},\xs) 
    &~\leq~ \tfrac{1}{\eta} f(\xs,\eta \loss^\expertset_{1:T}) 
    &\text{for all $\ExActs_{1:T}$}.
  \end{align}
\end{lemma}
\begin{proof}
  Construct a log loss game in which at any time $t$ each expert
  $\expert$ predicts according to a distribution $\ExPred{\expert}_t$
  such that $\ExPred{\expert}_t(x_t) = \exp(-\eta \loss_t^\expert)$
  for the actual outcome $x_t$ and $\ExPred{\expert}_t$ is arbitrary on
  other outcomes such that $\sum_{x_t} \ExPred{\expert}_t(x_t) = 1$. In
  this game the log loss of \AlgAlg{} is
  \begin{equation*}
    -\log \Palg(\xs[1:T])
    = \sum_{t\in1:T} -\log
      \walg[\x_t]{t}{\xs[\pastt],\eta \loss^\expertset_\pastt}.
  \end{equation*}
  By $\eta$-mixability of $\loss$
  \begin{align} 
    \loss(\a^{\Alg^\eta_{\loss}}_{1:T},\xs[1:T]) 
      &~=~ \sum_{t\in 1:T}
        \loss\Big(\predfunction\big(%
          \walg{t}{\xs[\pastt], \eta \loss^\expertset_\pastt}\big),
          x_t\Big)\notag\\
      &~\leq~ \tfrac{1}{\eta} \sum_{t\in 1:T}
        - \log \walg[x_t]{t}{\xs[\pastt],\eta
          \loss^\expertset_{\pastt}}.
       \label{eqn:derivedloss}
  \end{align}
  Combining with \eqref{eqn:loglossbound} and \eqref{eqn:derivedloss}
  completes the proof.
\end{proof}

Algorithms that satisfy the requirements of the lemma include Bayes,
follow the (perturbed) leader, the forward algorithm, MPP and
\OurAlg{}\@. An algorithm that does not satisfy them is the last-step
minimax algorithm~\cite{takimoto2000}, because it takes into account
the experts' predictions on outcomes that do not occur.

In the literature it is common to construct algorithms for arbitrary
mixable losses and point out their probabilistic interpretation for the
special case of log
loss~\cite{haussler1998,HerbsterWarmuth1998,bousquet2002}. Instead, we
have proceeded the other way around: first we derived results for log
loss and then we showed that they generalise to other losses. This
allowed us to draw on concepts and results from probability theory like
conditional probabilities, HMMs and the forward algorithm, without
reproving them in a more general setting.

\lemref{lem:otherlosses} generalises results by Vovk~\cite{Vovk1999},
who shows that the most important loss bounds for Bayes with logarithmic
loss can actually also be derived for arbitrary mixable losses. Our
algorithm \AlgAlg{} plays a role similar to his APA algorithm.

\section{Discussion}

\paragraph{Relearning vs Continuing to Learn}
\corlref{crl:twopart.loss.bound} bounds the regret of \OurAlg{} with
respect to a reference partition $\Part$ by $-\log \BackRef(\Part)$.
Consider the asymptotic behaviour of this bound if $\Part$ has
infinitely many shifts. (A shift occurs when $\prev^\Part(t+1) \neq
t$.)
For both decaying past with $\gamma \le 1$ (e.g.\ following
recommendations in~\cite{bousquet2002}) and uniform past (see
\tabref{tab:mixing.schemes})  $\max_{0 \le j < t} \BackRef_{t+1}(j)$
goes to zero as a function of $t$. Thus, the cost per shift (be it to
continue an earlier cell or to start a new one) grows without bound. On
the other hand for fixed share $\BackRef_{t+1}(0) = \alpha$ for all $t$,
hence fixed share can start a new cell at fixed cost. It depends on the
structured expert whether continuing previously selected cells at
increasing cost is advantageous over relearning from scratch after each
shift at fixed cost. For EHMM experts with a finite state space $\Q$
(including Bayes), relearning from scratch will cost at most a factor
$\card{\Q}$ over learning on. This factor is constant, so that fixed
share will eventually win.

\section{Conclusion}\label{sec:discussion}
We revisited Freund's problem, which asks for a strategy for prediction with expert advice that suffers low additional loss compared to Freund's reference scheme. We discussed the solution by Bousquet and Warmuth, which interprets the experts as black boxes. We proposed a new interpretation of Freund's scheme which is natural for learning experts, namely to train experts on the subsequence on which they are evaluated. This allows the reference scheme to exploit local patterns in the data, and thus makes the problem harder.

We solved Freund's problem for structured experts that are represented as EHMMs, building on the work of Bousquet and Warmuth. We showed that our prediction strategies are efficient, and have desirable loss bounds that apply to all mixable losses.

\exclude{%
\paragraph{Uniform Past vs Decaying Past}
In~\cite{bousquet2002}, the authors herald decaying past as the mixing scheme with the best practical performance. But their results are due to specific parameter settings. Here is why. First suppose that all past posteriors $\PastPost_t$ have converged to the expert that is best for outcome $t$, and suppose that a switch is forced. Then uniform past predicts using the uniform mixture of past posteriors, and therefore it assigns probability to each expert equal to this expert's empirical frequency of being best. Thus, uniform past \emph{learns} the expert frequencies.
Decaying past favours later posteriors, and this bias may lead to extra loss.

In practice, not all posteriors have converged. 
Posteriors around forced switches are quite vague, and earlier posteriors are vaguer than later ones, which suggests that we should choose $\gamma$ close but not equal to zero. (Decaying past reduces to Uniform past for $\gamma =0$.)
Our improved loss bound shows that practical performance may be much better than what Bousquet and Warmuth's bound suggests. Can we find rules of thumb to choose $\gamma$ to get good actual performance?

}

\appendix

\newcommand{\support}{\operatorname{Sp}}
\section{Running Times\label{app:runningtime}}

We compare the running times on $T$ outcomes of \OurAlg{} and the
forward algorithm, with respect to an arbitrary EHMM $\A$ with a countable
set of hidden states $\Q$. For simplicity we assume that the sets of experts
$\expertset$ and outcomes $\X$ are finite.

Let $\Qs[t]$ denote the hidden state of $\A$ at time $t$, and let
$\pinit$, $\ptf$ and $\ppf$ denote $\A$'s other components. Both
algorithms base their predictions on a distribution $\PredPost_t$ on
$\Qs[t]$ at time $t$, but differ in how they update $\PredPost_t$ after
observing $\xs[t]$. 
As the number of computations for this step
depends on the size of the support of $\PredPost_t$ and on $\ptf$, we
will need the following concepts.
For any probability distribution $\pmass$ on $\Q$, let $\support(\pmass)
= \setc{\q \in \Q}{\pmass(\q) > 0}$ denote its support. We recursively
define $\Q_t$, the set of states reachable in exactly $t$ steps, and
$\Q_{\le t}$, the set of states reachable in at most $t$ steps, by
\begin{align*}
  \Q_1 &\df \support(\pinit),
  &
  \Q_{t+1} & \df \bigcup_{\q \in \Q_{t}} \!\support(\ptf[\q]),
  &
  \Q_{\le t} & \df \bigcup_{i \in 1:t} \!\Q_i.
\end{align*}
Obviously, $\Q_t \subseteq \Q_{\le t} \subseteq \Q$ holds for all $t$.
Let $\ntransitions(S) \df \sum_{q \in S} \card{\support(\ptf[q])}$ be the number of outgoing transitions from any set of states $S
\subseteq \Q$.

\subsection{Forward}
The forward algorithm computes $\PredPost_{t+1}$ by
conditioning $\PredPost_t$ on $x_t$ and applying the transition function
$\ptf$. As $\PredPost_t$ has support $\Q_t$, the forward algorithm
requires $O\del[\big]{\ntransitions(\Q_t)}$ work per time step, and $O\del[\big]{|\Q_t| + |\Q_{t+1}|}$ space. Notice that, for finite $\Q$, the number of
transitions is bounded by $\ntransitions(S) \leq |\Q|^2$ for any $S$. A
rough upper bound on the total running time of forward on $T$ outcomes
is therefore $O\del[\big]{|\Q|^2 T}$, which is linear in $T$.

\subsection{\OurAlgNotFancy{}}

The \OurAlg{} algorithm comes in two variants: one for sleeping and one
for freezing. For sleeping the order of the running time is determined
by the evolution of past posteriors (line~\ref{line:only.for.sleeping} in \algoref{alg:main}); for
freezing, which skips line~\ref{line:only.for.sleeping}, either computation of $\PredPost_t$
(line~\ref{line:lambda}) or of the next posterior (line~\ref{line:loss.update}) is the dominant step. The
main difference for the running times of the two variants, however, is
that in sleeping $\PastPost_j$ has support $\Q_t$ at any time $t$,
whereas for freezing $\PastPost_j$ has support $\Q_{\leq j}$.

\subsubsection{Uniform Past}

For the uniform past mixing scheme, one can keep track of $\sum_{j=0}^t
\PastPost_j(\qs[t])$ to speed up computation of $\PredPost_{t+1}$.

\paragraph{Sleeping} This even works for sleeping, because applying the
state evolution to this sum in line~\ref{line:only.for.sleeping} of the algorithm is equivalent to
applying it to the individual $\PastPost_j$ and then summing. Consequently, sleeping
requires $O\del[\big]{\ntransitions(\Q_t)}$ work and $O\del[\big]{|\Q_t| + |\Q_{t+1}|}$
space per time step, which makes it as efficient as the forward
algorithm.

\paragraph{Freezing} For freezing, computing the next posterior (line~\ref{line:loss.update})
determines the running time. It requires $O\del[\big]{\ntransitions(\Q_{\leq
t})}$ work and $O\del[\big]{|\Q_{\leq t+1}|}$ space per time
step. Depending on the EHMM $\A$, this may be significantly slower than
the forward algorithm. First, 
for finite $\Q$, each of $\Q_t$, $\Q_{\leq t}$ and $ \Q$ have size $O(1)$ in $t$, and freezing runs in time $O(T)$, just like the forward algorithm.
Second, for infinite $\Q$, $\Q_{\leq t}$ may be unbounded as a function of $t$. Still, on $T$ outcomes
  \begin{equation*}
    \sum_{t\in 1:T} \ntransitions(\Q_{\leq t})
      ~\leq~ T \ntransitions(\Q_{\leq T})
      ~\leq~ T \sum_{t\in 1:T} \ntransitions(\Q_t),
  \end{equation*}
  which implies that freezing is no more than a factor $T$ slower than
  the forward algorithm.

\subsubsection{Decaying Past}

For the decaying past scheme the relative mixing weights of any two past
posteriors change from $\BackRef_t$ to $\BackRef_{t+1}$, which prevents
us from summing them as for uniform past. Implementing decaying past
therefore slows down both the evolution of past posteriors and
computation of $\PredPost_t$ by a factor of $O(t)$, and increases the
required space by the same factor. Fortunately, however, the decaying
past scheme can be approximated using a logarithmic number of uniform
blocks, as described in Appendix~C of~\cite{bousquet2002}. This reduces
the slowdown factor from $O(t)$ to $O(\log
t)$.\footnote{In~\cite{bousquet2002} it is suggested to weight each
block of posteriors $\PastPost_{[j_1,j_2-1]}$ by $(j_2 -
j_1)\BackRef_t(j_1)$. It seems that a marginal improvement is possible
by weighting by $\sum_{j_1 \leq j < j_2} \BackRef_t(j)$ instead, which
can be implemented equally efficiently for decaying past.} Thus, both
for sleeping and for freezing, approximated decaying past is only a
factor $O(\log T)$ slower than uniform past on $T$ outcomes, and
requires only a factor $O(\log T)$ more space.

\section{Loss Bounds}\label{app:loss.bound.proof}

We identify $\PredPost_t$ with the EHMM on $\family{\Qs[i], \Xis[i], \Xs[i]}{i \ge t}$ with initial distribution $\PredPost_t$, and with the transition and production functions of $\A$. So in particular $\PredPost_1 = \A$. For convenience, we shorten $\del{\PredPost_t}^\fr_{\Cell}(\xs[\Cell])$ to $\PredPost_t^\fr(\xs[\Cell])$ and $\del{\PredPost_t}^\sl_{\Cell}(\xs[\Cell])$ to $\PredPost_t^\sl(\xs[\Cell])$. Thus, among others, $\PredPost_t(x_t) = \PredPost_t^\sl(x_t) = \PredPost_t^\fr(x_t)$. 

\begin{lemma}\label{lem:backport}
For any $\Cell \subseteq t:T$, interpreting $\PredPost_0(\cdot|\xs[0])$ as $\PredPost_1$,
\begin{align*} 
\PredPost^\fos_t(\xs[\Cell]) 
&= \sum_{j \in 0:t-1} \beta_t(j) \PredPost^\fos_{j}\delc{\xs[\Cell]}{\xs[j]}.
\end{align*}
\end{lemma}
\begin{proof}
Let $\PastPost^t_j$ denote the past posterior $\PastPost_j$ at the beginning of round $t$. Thus for freezing $\PastPost^t_j = \PastPost_j$, and for sleeping $\PastPost^t_j$ is $\PastPost_j$ evolved $t-j$ steps. Then by definition $\PredPost_t(\xs[\Cell]) = \sum_{j= 0}^{t-1} \beta_t(j) \PastPost^t_{j+1}\del{\xs[\Cell]}$.
The operations $(\cdot)^\fr$ and $(\cdot)^\sl$ distribute over taking mixtures. The lemma follows from the fact that
$
\del{\PastPost^t_j}^\sl(\xs[\Cell]) 
=
\PastPost_j^\sl(\xs[\Cell]) 
$ and $
\del{\PastPost^t_j}^\fr(\xs[\Cell]) 
=
\PastPost_j^\fr(\xs[\Cell]) 
$.
\end{proof}

\newcommand{\somevar}{k}
\begin{proof}[Proof of \thmref{thm:loss.bound}]
  For any $t$, we view the mixing scheme $\BackRef_t$ as defining
  the distribution of a randomized choice $j_t \in 0:(t-1)$ for the
  predecessor of the $t$th outcome. 
  Let $j_{>\somevar} \df j_{\somevar+1:T} = (j_{\somevar+1}, \ldots,
  j_T)$ denote a vector of the choices beyond turn $\somevar$. Unfortunately,
  some choices of $j_{>\somevar}$ are inconsistent with any partition, because an
  element can only have one successor in a partition. Thus $j_{>\somevar}$ is
  inconsistent with any partition if $j_m = j_n > 0$ for $\somevar < m \neq n \le T$.
Let the predicate 
  $I(j_{>\somevar})$ be true iff $j_{>\somevar}$ is consistent with some
  partition.

  Some elements of $j_{>\somevar}$ may indicate the start of a new cell of the
  partition. Let $S(j_{>\somevar})$ denote the set of times when $j_{>\somevar}$ prescribes
  to start a new cell, i.e.\ $S(j_{>\somevar}) \df \setc{t \in \somevar+1:T}{j_t = 0}$.
  For an example, consult \figref{fig:not.ex}.

  Consistent values of $j_{>\somevar}$ specify the last part of a partition. For any $1
  \leq t \leq \somevar$, we may ask which of the times $\somevar+1:T$ will
  be put in the same cell as $t$. Let $R_t(j_{>\somevar})$ denote this set,
  including $t$.
For convenience, we abbreviate 
{%
\newcommand{\wideprod}[1]{\mathop{~~\prod~~}_{\!\!\!\!#1\!\!\!\!}}%
\begin{align*}
\BackRef(j_{>\somevar}) &\df \wideprod{t \in \somevar+1:T} \BackRef_t(j_t),
\\
W(j_{>\somevar}) &\df \wideprod{i \in S(j_{>\somevar})} \PredPost^\fos_1\del{\xs[R_i(j_{>\somevar})]}, &&\text{and}
\\
U_l(j_{>\somevar}) &\df \wideprod{i \in 1:l} \PredPost^\fos_i\del{\xs[R_i(j_{>\somevar})]} &&\text{for all $l \leq \somevar$},
\end{align*}
}%
\begin{figure}%
\caption[]{Notation example. $T=10$, $\somevar=4$, $j_{>\somevar} = (2,0,6,7,5,9)$, $S(j_{>\somevar}) = \set{6}$, $R_2(j_{>\somevar}) = \set{2,5,9,10}$.}\label{fig:not.ex}
\centering
\vskip 1.5em
$\xymatrix@C=1em{
0 &  
1 & 
2 & 
3 & 
4 & 
5 \ar@(ul,ur)[lll]& 
6 \ar@(dl,dr)[llllll]& 
7 \ar[l]& 
8 \ar[l]&
9 \ar@(ul,ur)[llll]& 
10 \ar[l]
}$
\vskip 1.5em
\end{figure}%
to name the intermediate debris arising from the incremental reduction of $\fosprob[\A](\xs)$. $W$-terms deal with cells that are completely specified by $j_{>\somevar}$, while $U$-terms keep track of the remaining partially specified cells.
  The proof proceeds by downward induction on $\somevar$, with induction hypothesis
  \begin{equation}
    \prod_{i\in 1:T} \PredPost_i(x_i)
      \geq \sum_{j_{>\somevar} \mathrel{:} I(j_{>\somevar})} \BackRef(j_{>\somevar}) W(j_{>\somevar}) U_\somevar(j_{>\somevar}).
    \label{eqn:inductionhypothesis}
  \end{equation}
For the base case $\somevar=T$ the hypothesis holds with equality, and
  for $\somevar=0$ the hypothesis is equivalent to the desired result \eqref{eqn:fosbound}.
It remains to verify that it holds for $\somevar-1 \ge 0$ if it holds for $\somevar$. To
  this end, fix $\somevar \ge 1$. To prove \eqref{eqn:inductionhypothesis}, it suffices to show that for consistent $j_{>\somevar}$
  \begin{equation*}
    W(j_{>\somevar}) U_\somevar(j_{>\somevar})
      \geq \sum_{j_\somevar \mathrel{:} I(j_{\ge\somevar})} {\kern-.5em}\BackRef_{\somevar}(j_\somevar)
        W(j_{\ge\somevar}) U_{\somevar-1}(j_{\ge\somevar}),
  \end{equation*}
  where $j_{\ge\somevar}$ denotes $j_{\somevar:T}$, i.e.\ $j_\somevar$ followed by $j_{>\somevar}$. 
  We expand the last factor of $U_\somevar(j_{>\somevar})$ using \lemref{lem:backport},  and bound
  \begin{align*}
    U_\somevar(j_{>\somevar})
      &= \sum_{j_\somevar\in 0:\somevar-1} \BackRef_{\somevar}(j_\somevar)
          \PredPost^\fos_{j_\somevar}\delc{\xs[R_{\somevar}(j_{>\somevar})]}{\xs[j_\somevar]} U_{\somevar-1}(j_{>\somevar})\\
      &\geq \sum_{j_\somevar \mathrel{:} I(j_{\ge\somevar})}\BackRef_{\somevar}(j_\somevar)
          \PredPost^\fos_{j_\somevar}\delc{\xs[R_{\somevar}(j_{>\somevar})]}{\xs[j_\somevar]} U_{\somevar-1}(j_{>\somevar}).
  \end{align*}
Observe that $R_t(j_{>\somevar}) = R_t(j_{\ge\somevar})$ for all $1 \le t < \somevar$ except $t=j_\somevar$. There are two cases.
If $j_\somevar=0$,
  then $U_{\somevar-1}(j_{>\somevar}) = U_{\somevar-1}(j_{\ge\somevar})$ and 
$W(j_{>\somevar})  \PredPost^\fos_1\del{\xs[R_{\somevar}(j_{>\somevar})]} = W(j_{\ge\somevar})$  
; 
on the other hand if $j_\somevar>0$ then 
$W(j_{>\somevar}) = W(j_{\ge\somevar})$. For consistent $j_{\ge\somevar}$, $U_{\somevar-1}(j_{>\somevar})$ contains the
  factor $\PredPost^\fos_{j_\somevar}(x_{j_\somevar})$, which implies that
  \[
     \PredPost^\fos_{j_\somevar}\delc{\xs[R_{\somevar}(j_{>\somevar})]}{x_{j_\somevar}} U_{\somevar-1}(j_{>\somevar})
      = U_{\somevar-1}(j_{\ge\somevar}). \qedhere
  \]
\end{proof}

\section{Invariance}\label{app:invariance}
\begin{proof}[Proof of \thmref{thm:invariance}]
Let $\mu^1$ and $\mu^2$ be distributions on $\Q^1$ and $\Q^2$. We overload notation, and write $\mu^1$ and $\mu^2$ for the EHMMs $\A^1$ and $\A^2$ with initial distribution replaced by $\mu^1$ and $\mu^2$. Recall that  $\mu^1$ and $\mu^2$ are equivalent if $\mu^1(\es) = \mu^2(\es)$ for all $\es$. Thus, $\A^1$ and $\A^2$ are equivalent iff $\pinit^1$ and $\pinit^2$ are equivalent.

To prove the theorem, we need to prove that equivalence is preserved by all the operations that \OurAlg{} performs, i.e.\ taking mixtures, performing loss update and performing state evolution.
Mixtures of equivalent distributions are equivalent, since  mixing and marginalisation commute. For loss update, note that $\ExPred{\e_1}_1(\xs[1]) = \mu^1(\xs[1]|\es) = \mu^2(\xs[1]|\es)$ for all $\ExPred{\E}_1$ and all $\es$. Finally, for state evolution, the claim follows from
\(
(\ptf{} \mathbin{\circ} \mu)(\es) = \mu(\Es[2:T+1] = \es)
\).
\end{proof}

{\footnotesize \bibliography{writeup,experts}}

\end{document}